\documentclass[10pt,twocolumn,letterpaper]{article}

\usepackage[breaklinks=true,bookmarks=false]{hyperref}
\usepackage{cvpr}
\usepackage{times}
\usepackage{epsfig}
\usepackage{graphicx}
\usepackage{amsmath}
\usepackage{amssymb}
\usepackage{balance}
\usepackage{booktabs}
\usepackage{lipsum}

\usepackage{capt-of,etoolbox} 

\cvprfinalcopy 

\def\cvprPaperID{19} 

\ifcvprfinal\fi
\begin{document}

\makeatletter



\title{Structured Query-Based Image Retrieval Using Scene Graphs}

\author{Brigit Schroeder\\
University of California, Santa Cruz\\
{\tt\small brschroe@ucsc.edu}
\and
Subarna Tripathi\\
Intel Labs\\
{\tt\small subarna.tripathi@intel.com}
}

\maketitle

\begin{abstract}
   A structured query can capture the complexity of object interactions 
   (e.g. 'woman rides motorcycle') 
   unlike single objects (e.g. 'woman' or 'motorcycle'). 
   Retrieval using structured queries therefore is much more useful than single object retrieval, but a much more challenging problem. 
   In this paper we present a method which uses scene graph embeddings as the basis for an approach to image retrieval. We examine how visual relationships, derived from scene graphs, can be used as structured queries. The visual relationships are directed subgraphs of the scene graph with a subject and object as nodes connected by a predicate relationship. Notably, we are able to achieve high recall even on low to medium frequency objects found in the long-tailed COCO-Stuff dataset, and find that adding a visual relationship-inspired loss boosts our recall by  10\% in the best case.
\end{abstract}

\ifcvprfinal\thispagestyle{empty}\fi

\section{Introduction}

An image is composed of a complex arrangement of objects and their relationships to each other. As noted in \cite{DBLP:conf/eccv/LanY0M12}\cite{Johnson_2015_CVPR}, this is why content-based image retrieval is more successful when using complex structured queries (e.g. `girl programs computer') rather than simply using single object instances 
(e.g. 'girl', 'computer', etc.). Instead of viewing objects in isolation, they can be coupled as a subject and object by a relationship that describes their interaction \cite{HuCCZG:19}. These \emph{visual relationships}, in the form of \textit{$<$subject, predicate, object$>$}  (e.g. \textit{$<$girl, programs, computer$>$}), can be used 
as
complex structured queries \cite{DBLP:conf/eccv/LanY0M12}  for performing image retrieval, which are 
more descriptive and efficient via their specificity.  

\begin{figure}[ht]
\begin{center}
\includegraphics[width=0.4\textwidth]{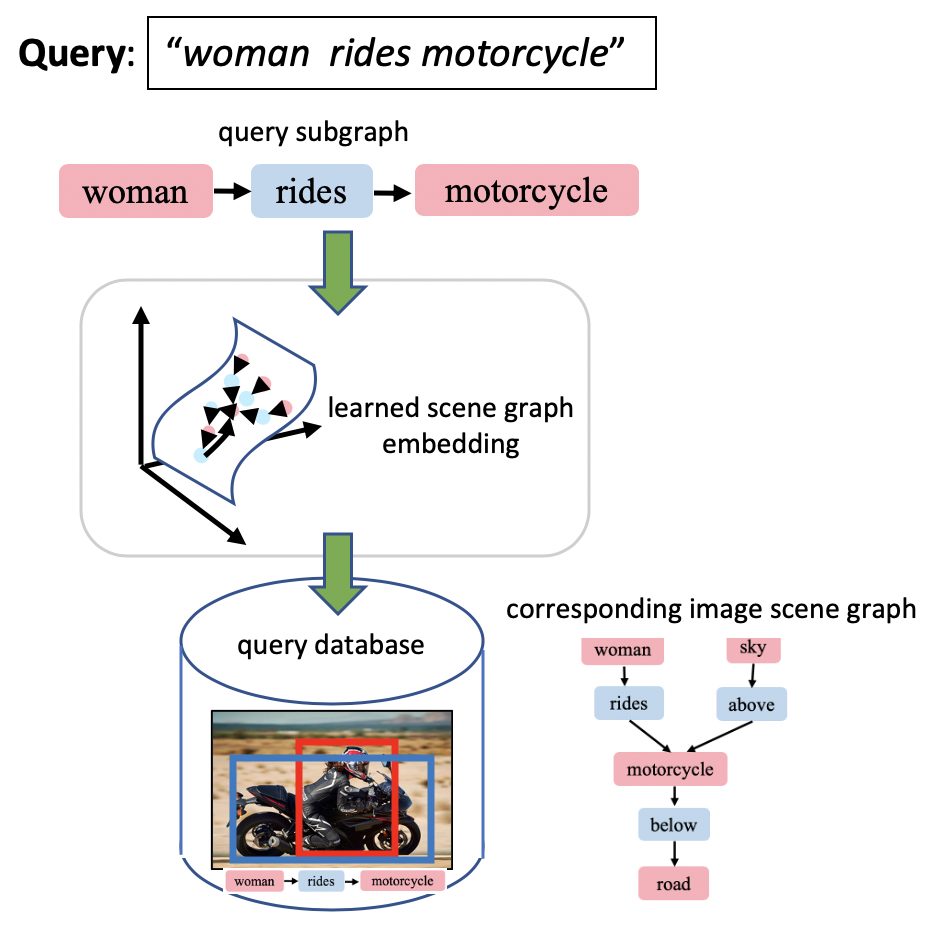}
\caption{\textbf{Visual Relationship Subgraph Query.} We use visual relationships, represented as directed subgraphs extracted from scene graphs, to form structured queries.  Each subgraph contains a subject and object as nodes connected by an edge representing a predicate relationship.}
\label{fig:subgraph_retrieval}
\end{center}
\end{figure}

\raggedbottom
\indent Visual relation-based 
retrieval, 
although more useful, is much more difficult than single object-based retrieval 
as the representation learning for the former case is more difficult.  
For example, the current state-of-the-art (SOTA) for object detection including both small object and long-tailed distributions achieves well over 50\% \cite{coco-detection-leaderboard} mean average precision (mAP) and recall at k=100 is about 70\%. However, when it comes to visual relationship detection (in particular `scene graph detection task'), even only for large and frequent objects from Visual Genome, the SOTA \cite{Zhang_2019_CVPR} recall at k=100 
is below 35\%.  This intuitively speaks to the difficulty of working with visual relationship representations, and thus the retrieval task at hand.

\begin{figure*}[ht]
\begin{center}
\includegraphics[width=0.8\textwidth]{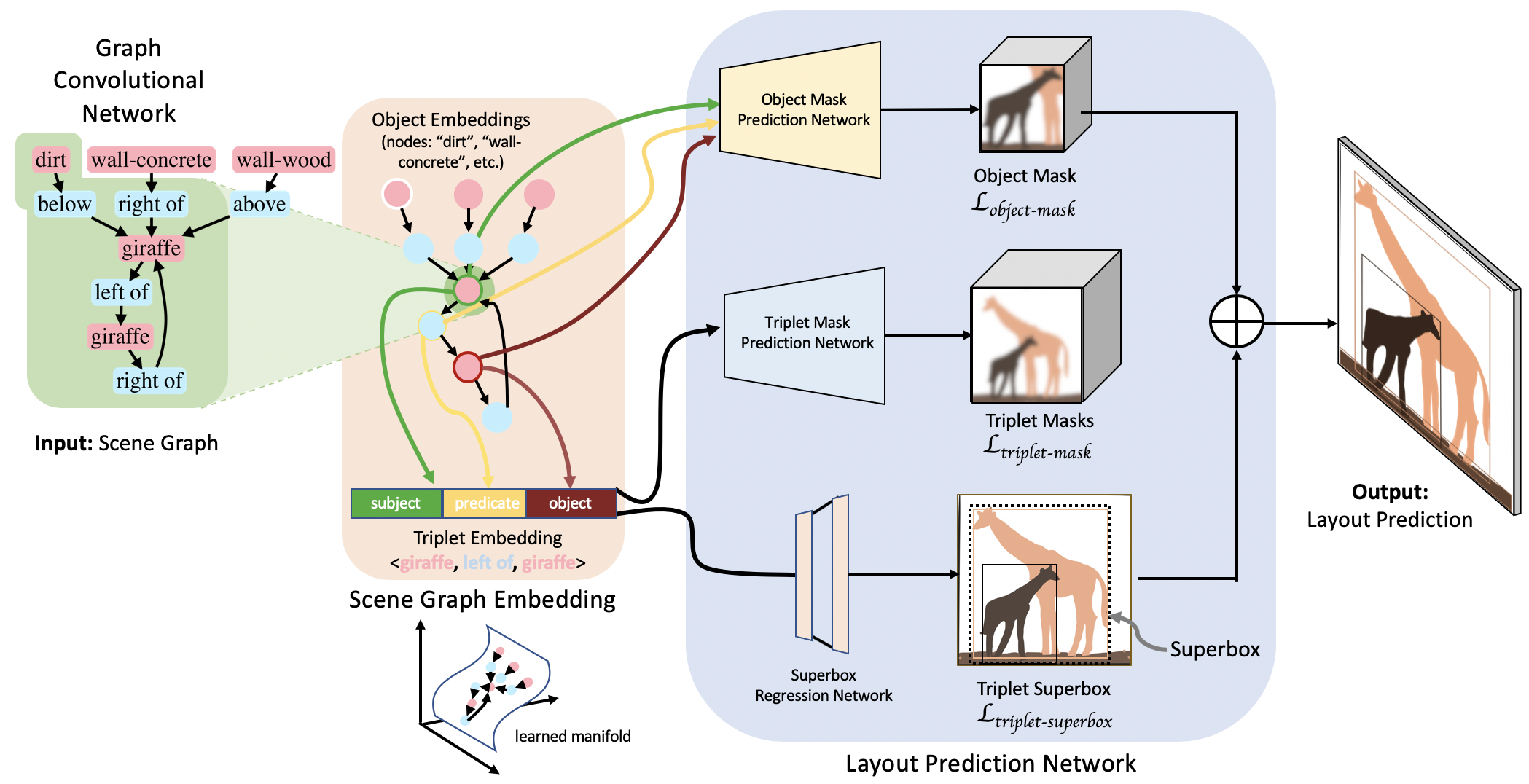}
\caption{\textbf{Scene Graph Embeddings from Layout Prediction.} A scene graph embedding is learned 
via a \textit{pretext task} which is 
training a layout prediction network. 
Later, 
image retrieval from structured queries, 
the downstream application we aim to address,
uses the 
similarity metric 
in the learned scene graph embedding space.}
\label{fig:sg_embedding}
\end{center}
\end{figure*}

In this paper, we approach the 
retrieval problem \ref{fig:subgraph_retrieval} by using a learned scene graph embedding from a scene layout prediction model (Fig  \ref{fig:sg_embedding}). Scene graphs are a structured data format which encodes semantic relationships between objects \cite{xu2017scenegraph}. Objects are represented as nodes in the graph and are connected by edges that express relationship, in the form of triplets. 
As shown in Figure \ref{fig:subgraph_retrieval}, 
we use visual relationships, represented as directed subgraphs extracted from scene graphs, to form structured queries.  Each subgraph contains a subject and object as nodes connected by an edge representing a predicate relationship.

Our work is unique in that we perform retrieval using only the embeddings extracted from scene graphs rather than visual features, which is a common modality in image retrieval \cite{Johnson_2015_CVPR}. We perform a quantitative and qualitative analysis which demonstrates our method's efficacy when dealing with a long-tailed dataset with overwhelming majority of low-frequency classes. We also observe that learning objectives derived directly from the visual relationships boost the image retrieval efficiency significantly.


\section{Related Work}
\indent
Early work by Johnson \etal \cite{Johnson_2015_CVPR} uses visual relationships derived from scene graphs for image retrieval. A scene graph is a structured data format which encodes semantic relationships between objects. A set of visual relationships containing a \textit{$<$subject, predicate, object$>$} are the building blocks of these scene graphs. 
Johnson \etal \cite{Johnson_2015_CVPR} use a conditional random field (CRF) model trained over the distribution of object groundings (bounding boxes) contained in the annotated scene graphs. 
Object classification  models are used as part of the CRF formulation. 
Wang \etal \cite{Wang_2020_WACV} use cross-modal scene graphs for image-text retrieval where they rely upon using word embeddings and image features. 
We distinguish our approach from \cite{Wang_2020_WACV} \cite{Johnson_2015_CVPR}  
as we do not use object groundings, word embeddings nor the input image features to perform our retrieval. 

A line of work that emerged recently \cite{img_gen_from_scene_graph18, ICLR_LLD, interactiveGenICLR19, Tripathi2019CompactSG, Jyothi2019LayoutVAESS, ICCV_interactiveGen19, vo2019visualrelation}
takes scene graphs as input and produce final RGB images. All of these methods perform an intermediate layout prediction by learning embeddings of nodes. We use layout generation as a pretext task for learning the embedding to perform image retrieval as the downstream application. 
However, unlike the above layout generation models, our method utilizes \textit{$<$subject, predicate, object$>$} triplets as additional supervisory signal for more effective structured prediction \cite{Schroeder_2019_ICCV}.
In  a closely  related  work,  Belilovsky \etal \cite{belilovsky:hal-01667777}  learn a joint visual-scene graph embedding for use in image retrieval. In  contrast,  our model is trained from scene graphs without access to visual features.

\section{Method}

\subsection{Dataset}


In this work, we use the 2017 COCO-Stuff \cite{cocostuff_caesar2018cvpr} dataset to generate synthetic scene graphs with clean predicate annotations. COCO-Stuff augments  
the  COCO dataset \cite{Lin2014MicrosoftCC} with additional stuff categories. The dataset annotates 40K train and 5K val images with bounding boxes and  segmentation  masks  for 80 thing categories  (people, cars, etc.)  and 91 stuff categories (sky, grass, etc.).  Similar to \cite{img_gen_from_scene_graph18}, we used thing and stuff annotations to construct synthetic scene graphs based on the 2D image coordinates of the objects. Six mutually exclusive geometric relationships are encoded and used as the predicate in visual relationships:  \emph{left of, right of, above, below, inside, surrounding}.

\begin{figure}[ht]
\begin{center}
\includegraphics[width=0.5\textwidth]{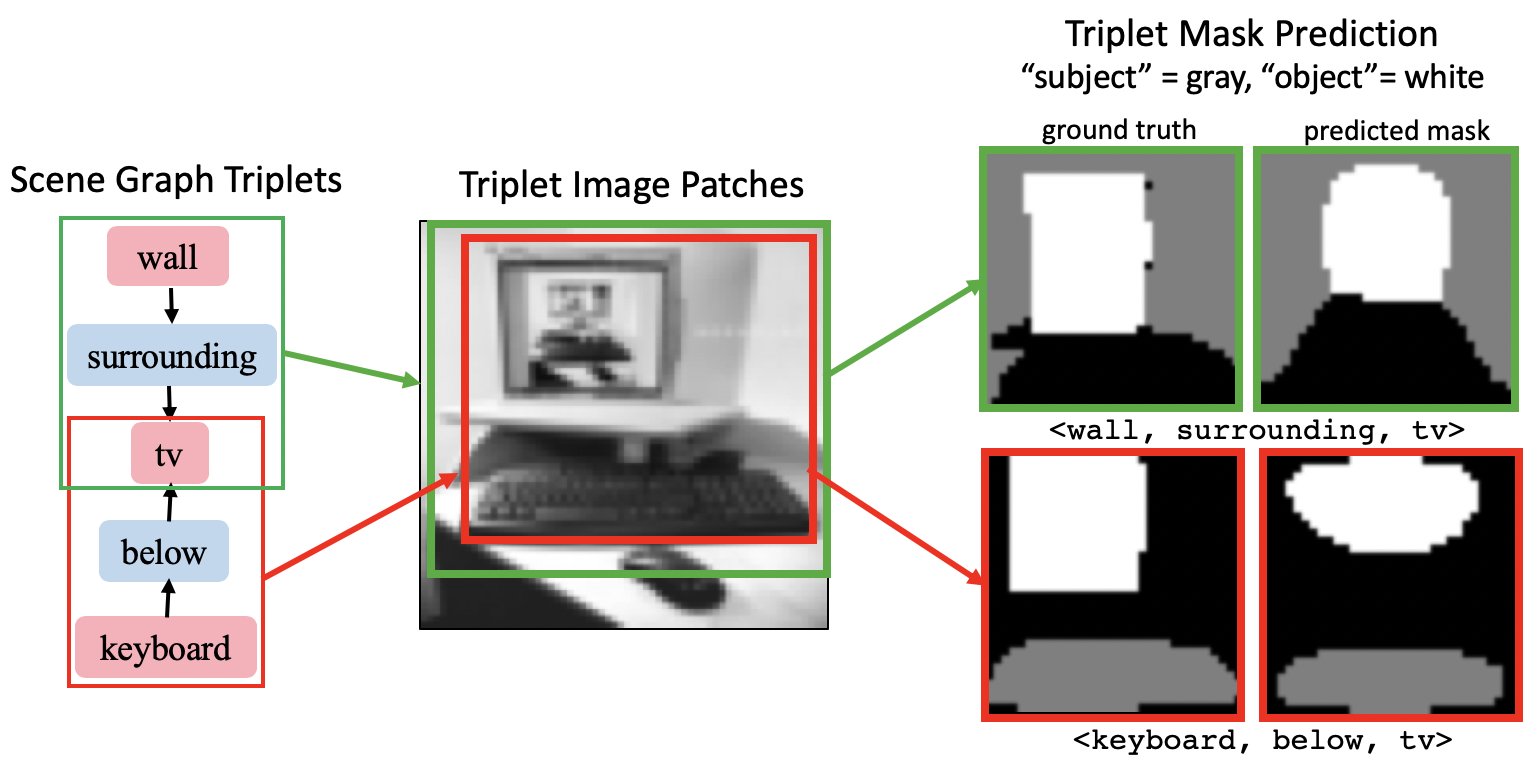}
\caption{\textbf{Triplet Mask Prediction.} Triplets containing a \textit{$<$subject,predicate,object$>$} found in a scene graph are used to predict corresponding triplet masks, labelling pixels either as subject and object. The mask prediction is used as supervisory signal during training.}
\label{fig:triplet_diagram}
\end{center}
\end{figure}

\begin{figure*}[ht]
\begin{center}
\includegraphics[width=1.0\textwidth]{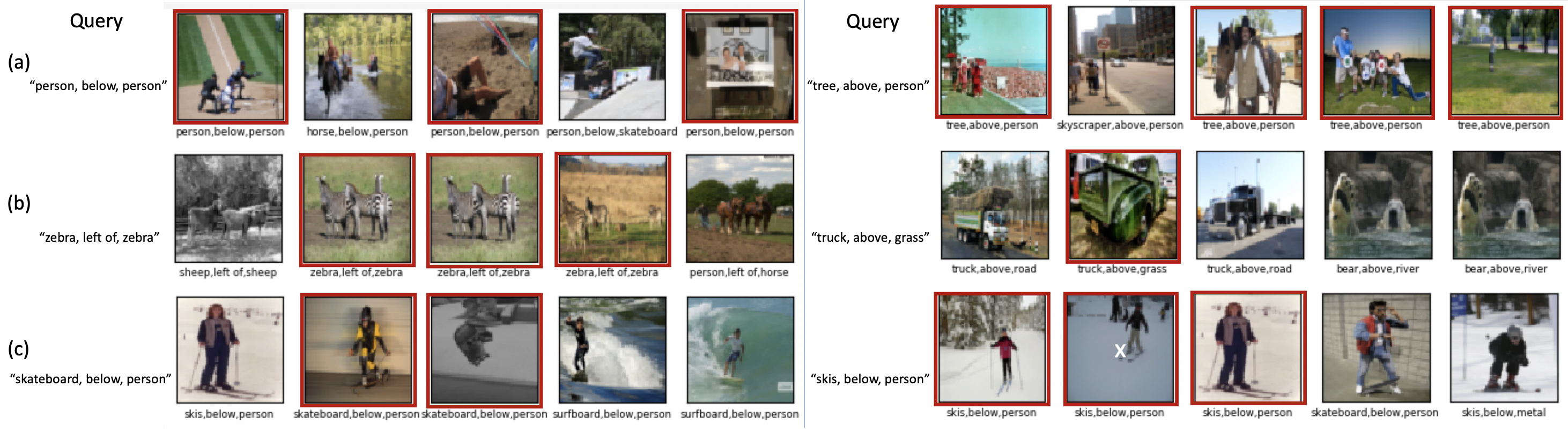}
\caption{\textbf{Image Retrieval Results.} Retrieval for structured queries with object types with varying levels of frequency in COCO-Stuff dataset: (a) head (\emph{person, tree}), (b) (long-tail) medium frequency (\emph{zebra, truck}), and (c) (long-tail) low frequency (\emph{skateboard, skis}). Query is in left-most column corresponding to red boxes.}
\label{fig:img_retrieval}
\end{center}
\end{figure*}
\subsection{Learning a Scene Graph Embedding}

We use a layout prediction network \cite{Schroeder_2019_ICCV} inspired by the image generation pipeline in \cite{img_gen_from_scene_graph18} to learn a scene graph embedding for image retrieval. 
Figure \ref{fig:sg_embedding} gives an overview of the network architecture. A graph convolutional neural network (GCNN) processes an input scene graph to produce embeddings corresponding to object nodes in the graph. The GCNN is a 5-layer multilayer perceptron where $D_{input}$ =  $D_{output}$ = 128 and $D_{hidden}$ = 512. Singleton object 
embeddings are passed to the next stage of the layout prediction network per \cite{img_gen_from_scene_graph18}. The outputs of the second stage of the layout prediction model are used to compose a scene layout mask with object localization. 
We utilize the object embeddings to form a set of triplet embeddings where each is composed of a \textit{$<$subject, predicate, object$>$}. 
We pass these through a triplet mask prediction network which learns to label objects as either `subject' or ’object' (see Figure \ref{fig:triplet_diagram}), enforcing both an ordering and relationship between objects.  We also pass triplet embeddings through a triplet `superbox' regression network, where we train the network  for joint localization over subject and object bounding boxes. A superbox is defined as the enclosing bounding box of both the subject and object bounding boxes as noted in Figure \ref{fig:sg_embedding}.



In this work, the layout prediction can be thought of as a pretext task in which the scene graph embedding is learned as an intermediary feature representation.  To improve the learning in this task, we apply two triplet-based losses \cite{Schroeder_2019_ICCV}  in addition to those used in \cite{img_gen_from_scene_graph18}. 
The first is a triplet mask loss, $L_{triplet-mask}$, penalizing differences between ground truth triplet masks and predicted triplet masks with pixelwise cross-entropy loss. The second is a triplet superbox loss, $L_{triplet-superbox}$, penalizing the $L_2$ difference between the ground truth and predicted triplet superboxes. We also train a layout prediction model without triplet-based losses for comparison.

\begin{figure}[th]
\begin{center}
\includegraphics[width=0.5\textwidth]{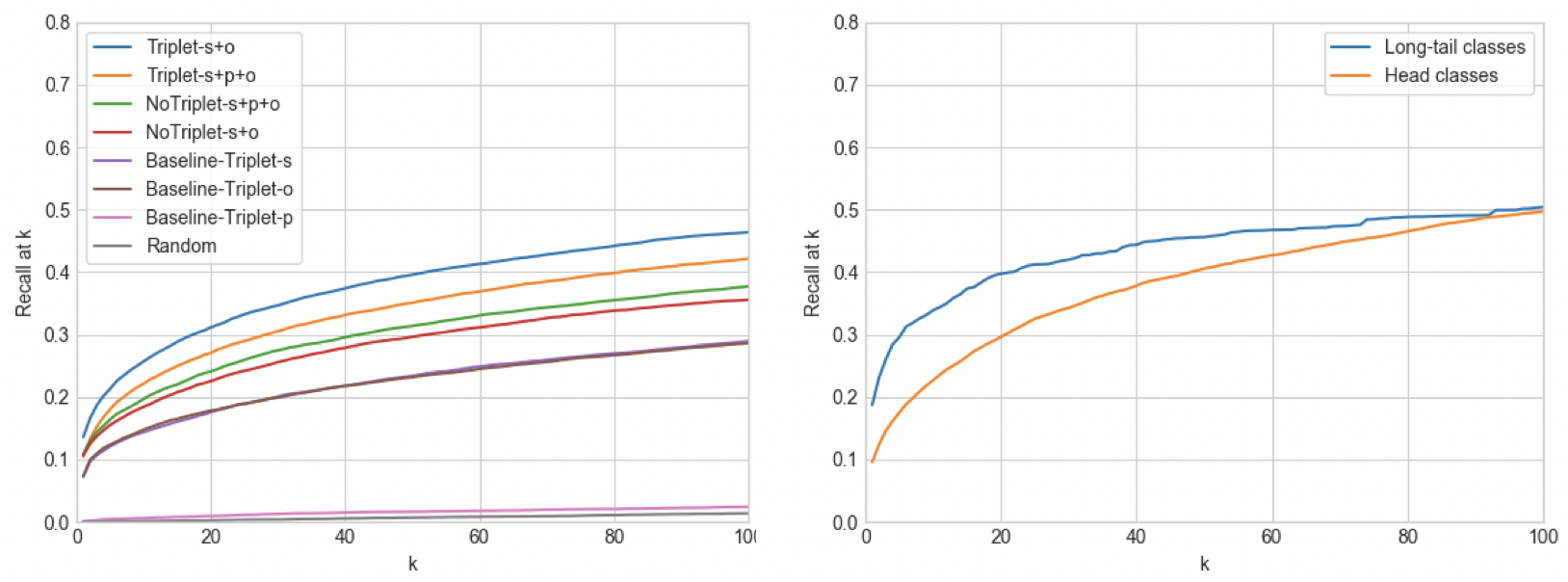}
\caption{\textbf{Image Retrieval Performance.} Recall@k for all classes (left) and long-tail vs. head classes (right) found in COCO-Stuff.}  
\label{fig:recall}
\end{center}
\end{figure}

\begin{table}[ht]
\begin{center}
\begin{tabular}{l|c|c|c|c} 
\toprule
Model & R@1 & R@25 & R@50 & R@100 \\ 
\midrule
{Triplet-s+o} & \textbf{0.14} & \textbf{0.33}  & \textbf{0.40} & \textbf{0.46} \\
{Triplet-s+p+o} & 0.10 & 0.29  & 0.35 & 0.42 \\
{NoTriplet-s+o} & 0.10 & 0.24  & 0.31 & 0.36 \\ 
{NoTriplet-s+p+o} & 0.11 & 0.26  & 0.31 & 0.38 \\
\hline
{Baseline-s} & 0.07 & 0.19  & 0.23 & 0.29 \\ 
{Baseline-o} & 0.07 & 0.19  & 0.23 & 0.29 \\ 
{Baseline-p} & 0.00 & 0.01  & 0.02 & 0.03 \\ 
{Random} & 0.00 & 0.00  & 0.01 & 0.01 \\ 
\midrule
\midrule
{Head Classes} &0.10 & 0.33  & 0.41 & 0.50 \\
{Long-tail Classes} & \textbf{0.19} & \textbf{0.41}  & \textbf{0.46} & \textbf{0.51} \\ 
\bottomrule
\end{tabular}
\end{center}
\label{table:layout_results}
\caption{\textbf{Recall@k.} Image retrieval performance is measured in terms of Recall@\{1,25,50,100\} for all classes (upper portion) and then separately for head and long-tail classes (bottom portion)}.
\end{table}


\section{Experimental Analysis}



\textbf{Query Database.} In this approach we focus on utilizing the object embeddings from a learned scene graph embedding to form structured queries. We have experimented with 
multiple forms of queries, including but not limited to visual relationships. For our testing, we query a database of 3100 visual relationships extracted from annotated test scene graphs from the COCO-Stuff dataset. We do not limit our database by requiring visual relationships to have a minimum number of occurrences as was done in \cite{Johnson_2015_CVPR}. We use 
the 
similarity metric $S$ to rank our retrieved 
images 
corresponding to their respective embedding space representation

\begin{equation}
S = \frac{1}{d(q,r_k)}
\end{equation}

where $d$ is the $L_2$ distance between the query $q$ and retrieved result $r_k$ at position $k$.


\begin{figure*}[ht]
  \includegraphics[width=\textwidth,height=4cm]{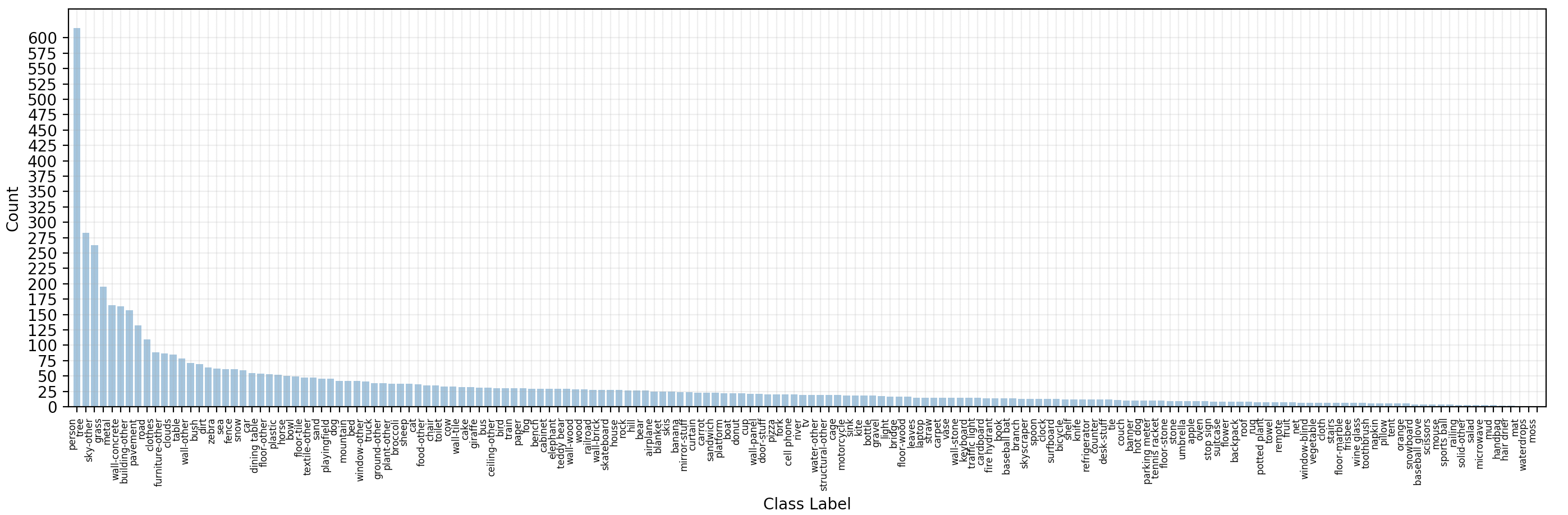}
  \caption{\textbf{Long-Tail Class Distribution.}  COCO-Stuff dataset has a long-tail object class distribution. This can be partitioned into head classes (first 20\%) and long-tailed classes (last 80\%).}
  \label{fig:coco_longtail}
\end{figure*}

\textbf{Long-tail Distributions.} We do not restrict the vocabulary of object classes which are used to form our queries as done in \cite{Johnson_2015_CVPR}. Given the long-tailed nature of the COCO-Stuff dataset (see Figure \ref{fig:coco_longtail}), it is important to acknowledge the difficulty a model may have in learning all classes sufficiently \cite{relation_prop_net}, especially for low frequency classes. We wish to understand how well our scene graph embedding performs given these challenges. Therefore, we divide objects classes into two parts for our experimental analysis. The first is head classes which comprise the first 20\% of dataset) (e.g.\emph{person}, \emph{tree} and \emph{sky}). The second is long-tail classes which comprise the remaining 80\% of dataset (e.g. \emph{zebra} , \emph{skateboard} and \emph{laptop}).  


\textbf{Results.} Initially, we break down our queries individually by subject (s), object (o) or predicate (p) as baselines. We see that the contribution from the subject and object embedding is much more significant over the predicate, as seen in Figure \ref{fig:recall} on the left. Image retrieval using only the predicate embedding is poor in terms of Recall@100 of 3\%, no better than random which has Recall@100 at 1\%. However, the subject and object embeddings both have a Recall@100 of 29\%, an increase of 26\% over the predicate. The reason for this could be that the triplet supervision biases the scene graph embedding towards subject and object, pointing to the need for more supervisory signals for the predicate embedding.

Given the minimal contribution of the predicate, we examine structuring our queries with and without using the predicate (p) from the visual relationship. We compare them using two types of models, those trained with and without triplet supervision (`Triplet' and `NoTriplet' in Figure \ref{fig:recall}). We see in Table 1 that the model trained with triplet supervision using a query structured with only subject and object (`Triplet-s+o') outperforms all model and query types by 10\% in the best case (36\% (`NoTriplet-s+o') vs. 46\% (`Triplet-s+o') for Recall@100). 

The omission of the predicate in the non-triplet modes is nominally worse (`NoTriplet-s+o' vs. `NoTriplet-s+p+o' in Figure \ref{fig:recall}). However, we clearly see that the omission of the predicate in the triplet models (`Triplet-s+p+o' vs. `Triplet-s+o') improves recall by 4\% (42\% vs. 46\% for Recall@100 for ). This follows the trend seen in the baseline of subject and object-only queries outperforming the predicate-based queries. We also observe that visual relationship-based queries (and structured variations thereof) outperform single object queries by 17\% in the best case (29\% (subject or object) alone vs. 46\% (`Triplet-s+o') for Recall@100). The triplet-based losses emphasize the interaction between subject, object and predicate embeddings, and this may lend to the significant boost seen in retrieval done with structured queries.

Figure \ref{fig:recall} (right) demonstrates the average retrieval performance on long-tail and head classes in the COCO-Stuff dataset. A low occurrence of an object class corresponds to a low occurrence of visual relationships with this object, making the task of image retrieval for long-tail classes more challenging. The long-tailed distribution of COCO-Stuff can be seen in Figure \ref{fig:coco_longtail}, where the majority of object classes  in the long-tail have a frequency (count) of less than 25 instances. Despite this, the long-tail classes have a high recall@k, especially when k is less than or equal to 10. Figure \ref{fig:recall} (right) and Table 1 show that long-tail classes tend to do at least as well as the high-frequency head classes or some cases, much better, especially at low values of k.  This is exemplified in Figure \ref{fig:img_retrieval} where even the middle to low frequency long-tail classes have several matches in the top k=5.

Qualitative retrieval results can be seen in Figure \ref{fig:img_retrieval} using a triplet-based model. Even with middle to low frequency classes found in the long-tail distribution of COCO-Stuff, we have successful retrieval for k=5. Importantly, note that we are able to have exact matches \emph{despite} omitting the predicate (e.g. s+o), and also exact matches in  predicate (only) for all retrieval results despite its omission.  Even incorrect results have the correct predicate and often are semantically similar (e.g.`surfboard below person' vs. `skateboard below person').

\section{Conclusion}

We have trained scene graph embeddings for layout prediction with triplet-based loss functions. For the downstream application of image retrieval, we use structured queries formed using the learned embeddings instead of input image content. Our approach achieves high recall even on long-tail object classes.

\balance
{\small
\bibliographystyle{unsrt}
\bibliography{main}
}

\end{document}